%% file: paper.tex
\documentclass[conference]{IEEEtran}
\IEEEoverridecommandlockouts
\usepackage{cite}
\usepackage{amsmath,amssymb,amsfonts}
\usepackage{graphicx}
\usepackage{textcomp}
\usepackage{xcolor}
\usepackage{subcaption}
\usepackage{algorithm}
\usepackage{algpseudocode}
\usepackage{hyperref}
\hypersetup{
    unicode=false,          
    pdftoolbar=true,        
    pdfmenubar=true,        
    pdffitwindow=false,     
    pdftitle={Procedural Generation of Complex Roundabouts For Autonomous Vehicle Testing},    
    pdfauthor={Golam Md Muktadir},     
    pdfsubject={Research Paper},   
    pdfcreator={Golam Md Muktadir},   
    pdfproducer={},  
    pdfnewwindow=true,      
    colorlinks=false,       
    linkcolor=red,          
    citecolor=green,        
    filecolor=magenta,      
    urlcolor=cyan           
}

\def\BibTeX{{\rm B\kern-.05em{\sc i\kern-.025em b}\kern-.08em
    T\kern-.1667em\lower.7ex\hbox{E}\kern-.125emX}}
\begin{document}

\title{
Procedural Generation of Complex Roundabouts for Autonomous Vehicle Testing
}
\author{
    
    
    \IEEEauthorblockN{
        Zarif Ikram
    }
    
    \IEEEauthorblockA{
       Computer Science and Engineering\\
        Bangladesh University of\\ Engineering and Technology, Dhaka\\
        1905111@ugrad.cse.buet.ac.bd
    }
    \and
    \IEEEauthorblockN{
        Golam Md Muktadir
    }
    \IEEEauthorblockA{
        Computer Science and Engineering\\
        University of California\\
        Santa Cruz, California\\
        muktadir@ucsc.edu
    }
    \and
    \IEEEauthorblockN{
        Jim Whitehead
    }
    \IEEEauthorblockA{
        Computational Media\\
        University of California\\
        Santa Cruz, California\\
        ejw@ucsc.edu
    }
}

\maketitle

\begin{abstract}
High-definition roads are an essential component of realistic driving scenario simulation for autonomous vehicle testing. Roundabouts are one of the key road segments that have not been thoroughly investigated. Based on the geometric constraints of the nearby road structure, this work presents a novel method for procedurally building roundabouts. The suggested method can result in roundabout lanes that are not perfectly circular and resemble real-world roundabouts by allowing approaching roadways to be connected to a roundabout at any angle. One can easily incorporate the roundabout in their HD road generation process or use the standalone roundabouts in scenario-based testing of autonomous driving.
\end{abstract}


\section{Introduction}
High-definition (HD) maps consist of roads, lanes, signs, buildings, and other road objects, essential in creating a realistic world in a simulator for autonomous vehicles (AV). Using HD maps in simulation-based testing improves the generalizability of test results to the real world because HD maps are detailed, accurate, extensive, and photo-realistic, thereby permitting an AV to use multiple sensors to build a 3D world model. An essential aspect of HD maps is road geometries' overall realism and variation. Due to their complexity, roundabouts are often omitted from procedural road network generators, yet they do occur in the real world. In some cities, they are very common. Since there are roundabouts of various shapes and sizes \cite{roundabouts}, an AV needs to have access to a wide variety of roundabout shapes for thorough simulation-based testing.

Roundabouts in HD maps can be generated manually (e.g., using a tool like RoadRunner) or via extraction from digital maps like OpenStreetMap(OSM) \cite{openstreetmap}. Manual authoring of roundabouts is time-consuming and labor-intensive, typically limiting the number and variety of instances that are created. Roundabouts extracted from digital maps may require manual cleanup and are challenging to modify \cite{AutomaticOSM}. Procedural generation of roundabouts addresses the problems with manual and extractive approaches -- it is fast, scalable, and can produce realistic yet different outputs.  

While existing approaches produce roundabouts with fixed structures that cannot be adapted to surrounding road structures, the approach described in this paper takes the surrounding information as constraints and produces roundabouts consistent with these constraints. This ensures the lanes and direction of the approaching roads to the roundabout area mold the roundabout. In addition to flexibly producing roundabouts, the method outputs the roads in OpenDRIVE \cite{asamopendrive} format, a standard language to describe road geometries. OpenDRIVE roads can easily be imported into driving simulators such as CARLA \cite{carla}.

This study is motivated by the restrictive design of existing roundabout generation methods. Our first finding is that current approaches frequently only permit connections at right angles and do not permit approaching roads to be joined at other angles. Realistic roundabouts, however, permit approaching roads to be joined at a variety of angles (\textit{see Figure \ref{fig:comparsion}}). Our second observation is that the lanes inside a roundabout are not always perfectly circular. Driving is more difficult due to the inside lanes' changing curvature. Our major contributions target these findings and develop methods to address them. Moreover, our generator can also produce Turbo Roundabouts\cite{fortuijn2009turbo}, which are not seen in any other generator.

In this work, we formalize the method that can capture the variability of a roundabout's angles and semicircular shapes. We provide the expressive range analysis of our output.

\input{contents/related_work.tex}
\section{METHODS FOR ROUNDABOUT GENERATION}
In this section, we present our approach for creating two types of roundabouts: (a) classical roundabout, (b) turbo roundabout. 
\input{tables-figures/roundabout_classification.tex}

A roundabout is a set of roads of two types: circular roads and incident roads. Vehicles approach and leave the roundabout through one of the incident roads. Circular roads create the loop inside the roundabout and connect the incident roads. Furthermore, we define the \textit{incident road definition} as a four-value tuple of \textit{(position, heading, numLeftLanes, numRightLanes)}. Each incident road definition corresponds to the incident point where an incident road is connected to the roundabout area. Given a set of incident road definitions, a roundabout is generated in three phases. In the first phase, we find a circle using the incident points. In the second phase, we create the circular roads. In the third phase, we create incident roads and connect them to circular roads using connection roads.

In our work, we use the term classic roundabouts to describe standard multi-lane roundabouts. Turbo roundabouts refer to roundabouts with enhanced design for disallowing lane changing, which minimizes conflict points.

\input{methods/classic.tex}

\input{methods/turbo.tex}

\input{tables-figures/beautifulroundabouts.tex}
\input{tables-figures/beautifulturboroundabouts.tex}
\section{EVALUATION}
We first present a qualitative evaluation of our work. Classic and turbo roundabouts produced by Junction-Art, as seen in Fig. \ref{fig:classicroundabouts} and \ref{fig:turboroundabouts}, vary in the number of incident roads, incident road angles, radius, lanes, etc. In Fig \ref{fig:comparsion}, an example of a real life roundabout is presented. It is apparent that the roundabout contains different numbers of lanes for different incident roads, which can also be seen in the output from our generator. In Feature 2, we can see the variable radius of the circular roads. Our work can produce similar results as well. Finally, in the real-world example, incident roads can have different incident angles, which is also reflected in our output.

\input{tables-figures/comparison.tex}
\input{tables-figures/pointgen.tex}
\input{tables-figures/circlevariation.tex}

To show the diversity of the curvature captured by our generator, we present an expressive range analysis of our output. First we generate 20 \textit{n-way} random roundabouts. The method requires us to first select a random circle with a radius of 35 to 45 meters and then randomly pick $n$ points. Then we randomize their heading, and pass them into our generator (Fig. \ref{fig:roaddef}). 

Fig. \ref{fig:circlecomp} shows the summary of the semicircular shape of the 3-way roundabouts. The radii distributions of the 3, 4, and 5-way generated roundabouts are presented in Fig. \ref{fig:radiusDis}. As shown in Fig. \ref{fig:3-way}, it is apparent that most distributions are centralized towards the 12 to 25 meter region, which is due to the radius choice of the randomized road-definition generator. Besides, most distributions are platykurtic, showing the variability of the radius in roundabouts. The central tendency and the flat nature of the distribution also persists in the 4 and 5-way roundabouts (Fig. \ref{fig:4-way} \& \ref{fig:5-way}).
\input{tables-figures/radiusdist.tex}
\input{tables-figures/derivative.tex}

To further analyze the variability of radii, we present the distribution of the derivative of the radii for several 3-way roundabouts in Fig. \ref{fig:derivative}. The distributions are generally centered towards 0, often slightly skewed, and leptokurtic. Hence, it is apparent that generated roundabouts, while keeping a circular resemblance, change radius smoothly.
\input{tables-figures/3wayfixedradius.tex}
\input{tables-figures/fixed_roundabout_dis.tex}
Finally, to show that the variations are unaffected by input variety, We generate 30 3-way roundabouts with a fixed input. As shown in Fig. \ref{fig:fixedRadDist}, the distribution of radii for each roundabout is different while centered at the same position, showing that the roundabouts have variations in shape while keeping the same circular structure. We also see a similar outcome in Fig. \ref{fig:fixed_dist} and \ref{fig:fixed_heatmap}, which shows no notable bias in the derivative of radii regardless of a fixed input or randomized inputs.

\section{Future Work}
While variations in both roundabouts were thoroughly presented in this work, there are many variation in turbo roundabouts that still need to be addressed such as shape, number of spikes etc. Many roundabouts have slip roads that are adjacent to or even sometimes partially overlapping with the roundabout area. We left such designs for future scope.
There are also room to improve the distortion algorithm. Algorithms that can generate different shapes could more naturally model real-world distortion distributions.


\section{CONCLUSIONS}
This paper details how JunctionArt can generate Classic and Turbo roundabouts. These roundabouts can adapt to the incident roads and hence are pluggable into existing HD road networks. We formulated a novel approach to connect approaching roads from any angle to roundabout lanes without violating the OpenDRIVE rules for describing roads. The roundabout lanes can take distorted circular shapes creating a variation in curvature. We present an expressive range analysis of the shapes and curvatures. We look forward to the inclusion of greater numbers of more varied roundabouts in future AV simulation based testing scenarios.Í









\bibliographystyle{IEEEtran}
\bibliography{IEEEabrv,paper}

\end{document}

%% file: contents/related_work.tex
\section{RELATED WORK/Literature Review}


Road generation is important for many different purposes, ranging from video games to modeling systems of various complexity. Many exciting approaches can produce roads with different details. Usually, most approaches generate roads by creating a network of reference lines with predefined lane widths. PCG-based road generation generally facilitates virtual game worlds and procedural city generation  \cite{greuter2003real, parish2001procedural, lechner2006procedural, chen2008interactive, smelik2014survey}. There is another body of work that generates roads based on terrain, population density, etc. \cite{mccrae2008sketch, galin2011authoring, freiknecht2017survey, Ilangovan2009ProceduralCG, kelvin2020procedural, hartmann2017streetgan, guerin2017interactive}. While some approaches like \cite{chen2008interactive} can generate radial-patterned road networks, none of the approaches address the generation of roundabouts, particularly because the connection of terrain, cities, etc. does not require roundabouts.

However, for AV testing, the generation of detailed roundabouts is essential to capture the complexity of real-world road networks. Detailed roundabouts can be generated in two ways: (a) using real-world data or (b) using algorithmic PCG techniques not based on distributions in real-world data. The generation of roads, including roundabouts, can benefit from using real-world data such as GIS data, real images, etc. The main motivation behind using real-world data to construct road geometries is that it can capture the varieties and complexities of real roads. On the other hand, PCG-based approaches generate roads by carefully mimicking the complexities seen in real life and algorithmically describing the generation process. These approaches excel at large-scale variations and improved control through the use of parameters.

\textit{Using real-world data:} StreetGen, \cite{streetgen}, generates road geometries from road networks made of polylines with estimated roadway widths from GIS data and, therefore, can construct roundabouts from real-world examples. However, it lacks precision since its lacks configurable lane width and types and also faces difficulty approximating arcs as segments. \cite{despine2011realistic} describes a method for automating the generation of a 3D virtual environment from GIS data and GPS navigation data. From their work, it is evident that they can reproduce the roundabouts seen in the data, capturing the complexity and realism. However, the generated roundabouts must be encoded in the input data. \cite{highfidelity} proposes a novel method of constructing 3D roads from 2D GIS data using civil engineering design guidelines. Their work includes elevation profiles, smooth transitions between consecutive segments, lane markings, complex intersections, etc. While this approach can produce roundabouts with a great degree of detail, it is not sufficient to handle the mass production of HD roads because of insufficient data and, hence, human intervention is required for road validation. Other recent approaches such as \cite{mapGenerationExtractFeature} and \cite{AnalysisAndVariation} extract features from input maps and construct maps maintaining scenario diversity. While they produce intersections with varied diversity, neither address the generation of roundabouts.

\textit{Constructive generation:} these methods approach the road generation problem without relying on real-world data or human intervention, and use algorithms to procedurally generate road geometries. By focusing on the variety and diversity of the produced outputs, this method can be used to produce scalable road networks. However, in order to automatically generate roads that capture the characteristics seen in the real world, constructive methods require a sound knowledge of the domain. PGDrive builds a road network by selecting and connecting configurable pre-defined road segments including roundabouts \cite{li2020improving}. In PGDrive's roundabouts, only the radius of the roundabouts is configurable, the number of incident roads at the roundabouts is fixed at 4, and the angle at which incident roads can connect to the circular junction is also fixed at 90 degrees. Hence, PGDrive does not represent the variety observed in real-world roundabouts. Other PCG-based approaches, such as Pyodrx \cite{pyodrx} and ASfault \cite{ASFault} cannot generate roundabouts. AmbieGen, a recent work, can produce challenging road segments for the Lane Keeping Assist System (LKAS), but cannot produce roundabouts. \cite{PCGOSM} claims to generate roundabouts using PCG methods from OSM data, but the detailed methods and evaluation are not presented in the work. A more advanced approach \cite{cvprMAP} generates HD maps using a hierarchical graph to represent maps in a data-driven way, but does not specify anything about roundabouts.



%% file: tables-figures/roundabout_classification.tex
\begin{figure}[ht]
    \begin{subfigure}[b]{.5\linewidth}
      \centering
      \framebox{\includegraphics[width=.9\linewidth]{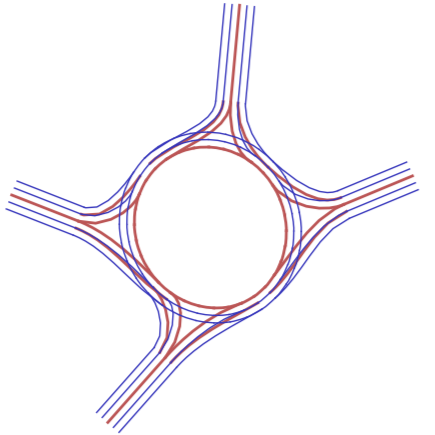}}
      \caption{}
      \label{fig:4-way-k}
    \end{subfigure}
    \begin{subfigure}[b]{.5\linewidth}
      \centering
      \framebox{\includegraphics[width=.9\linewidth]{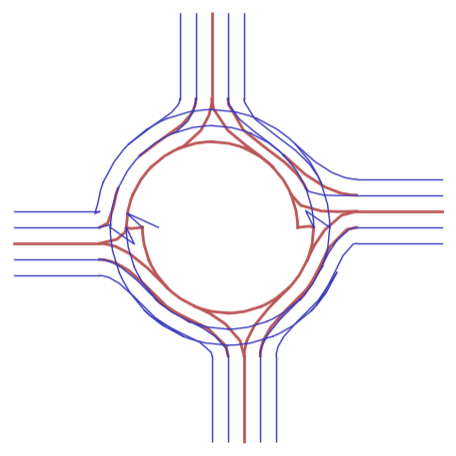}}
      \caption{}
    \end{subfigure}
    \caption{(a) Randomly generated classic roundabout with 4 incident roads, (b) Randomly generated turbo roundabout with 4 incident roads}
\end{figure}

%% file: methods/classic.tex
\subsection{Classic Roundabout}
This section discusses the three phases of classic roundabout generation.

\vspace{5mm}
\subsubsection{\textbf{Phase 1: Finding Maximal Circle Which Has No Incident Points Inside}}


\input{tables-figures/classic_generation.tex}

In phase 1 (Fig:\ref{fig:roundabout_phase-1}), we find a circle defined by $(C_x, C_y, r)$. Given incident points, we find a circle that contains no incident points inside. This is important as the circle lays the foundation for the circular roads, which cannot overlap with any incident roads. To find out the center, we use the least square regression method for circle fitting. Once the center $(C_x, C_y)$ is found, the radius can be found by finding the minimum distance between the center and the incident points.
\[r = 0.4 \min_{i = 1...n} distance((C_x, C_y), (x_i, y_i))\]

\vspace{5mm}
\subsubsection{\textbf{Phase 2: Creating Circular Roads}}
\input{tables-figures/phase2classic.tex}
We use the circle from phase 1 as the reference line of the circular road inside the roundabout. The creation of circular roads poses a significant challenge due to the specifications of ASAM OpenDRIVE. Circular roads can be built using only two half-circle segments. However, OpenDRIVE only permits roads to be linked to their start and end points\cite{opendrive1.6}. Hence, if incident roads are connected in the way shown in Fig:\ref{fig:invalidconnection}, while the connection is valid, the geometry becomes invalid due to road overlap. We first break down the circle into small segments to solve this challenge. Then we find their starting points and heading. Then we either create perfectly circular roads by joining the points with circular segments (Fig:\ref{fig:moresegments}) or, to induce an irregular circular shape, we add Perlin noise to the points and finally join the points using connection roads (Fig:\ref{fig:moresegmentPerlin}). A connection road is a parametric cubic road with lanes.





\vspace{5mm}
\subsubsection{\textbf{Phase 3: Creating Incident Roads}}


Phase 2 produces the body of a roundabout. Phase 3 extends incident roads from the incident points given by the input road definitions, so they join the roundabout (Fig:\ref{fig:roundabout_phase-3}). While connecting incident roads to circular roads becomes easier due to the modification is done in phase 2, deciding which point to connect to becomes difficult because incident angles can differ. Thus, a trivial rule, such as finding the closest circular segment for connecting straight roads, can create invalid geometries. For example, in Fig:\ref{fig:badconnection}, incident roads are connected to the right and left points of the closest point, and we can see that it causes connection roads to overlap with circular roads.

Therefore, we introduce \textit{centerOffset} in this phase (Fig.\ref{fig:offset}). \textit{centerOffset} is the angular difference between the original heading of an incident point and the direction of the vector between the center of the roundabout and the incident point. It is essential for making two important decisions:
\begin{itemize}
    \item Using \textit{centerOffset}, we can decide which circular road segment to connect a straight road to. If this method was not used, it would always connect straight roads to circular road segments based on their starting point, leading to undesirable connections.
    \item \textit{centerOffset} affects the length of a straight road. The variability of the straight roads widens the incident angle range of the generator.
\end{itemize}
Finally, phase 3 has the following steps:
\begin{enumerate}
    \item Create straight roads from road definitions based on distance from an incident point to the center and the \textit{centerOffset} of the incident point. The length of the road can be found by finding the smallest length for which the distance between the road endpoint and the center is minimum and the road does not overlap the circular roads. 
    \item Calculate road connection points for each roads using $centerOffset_{startingPoint}$ (Fig:\ref{fig:offset}).
    \item Join incident roads with their respective circular road segments.
\end{enumerate}
\input{tables-figures/offset_figure.tex}

%% file: tables-figures/classic_generation.tex
\begin{figure}[ht]
\begin{subfigure}[b]{.32\linewidth}
  \centering
  \framebox{\includegraphics[width=.9\linewidth]{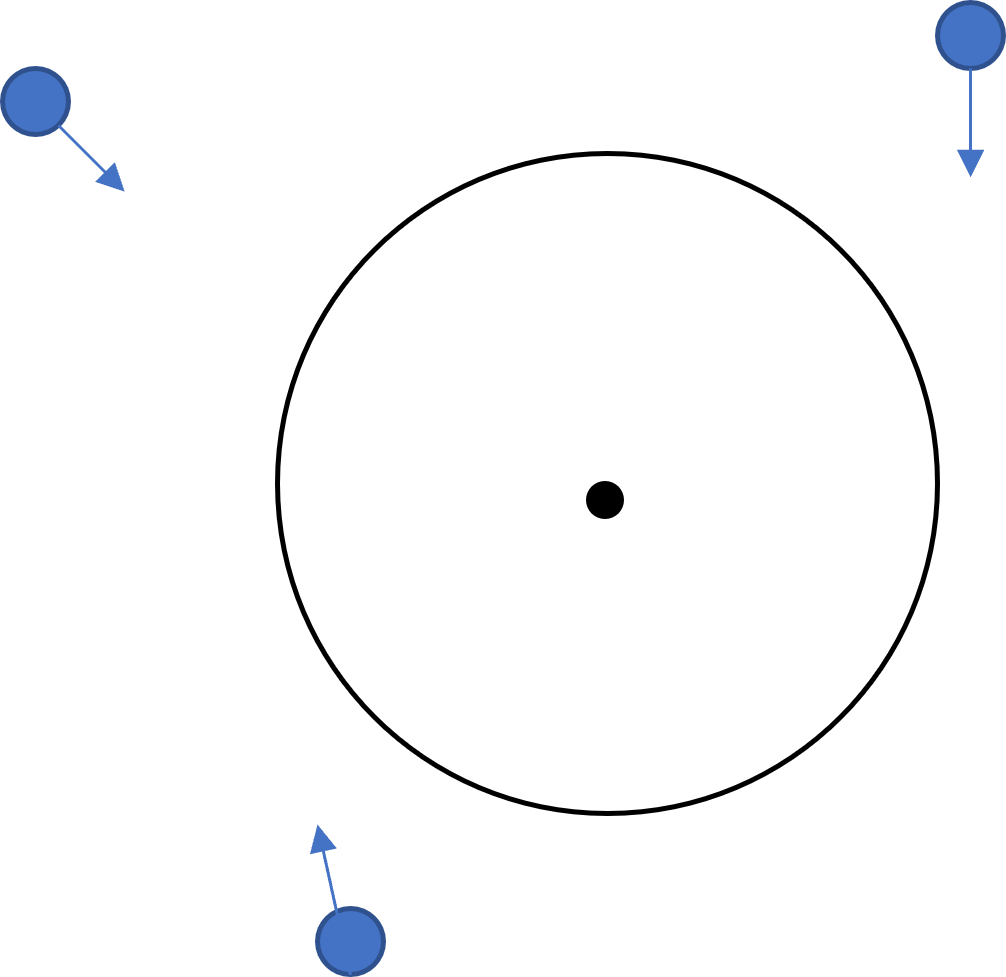}}
  \caption{circle}
  \label{fig:roundabout_phase-1}
\end{subfigure}\hspace{0.01\linewidth}%
\begin{subfigure}[b]{.32\linewidth}
  \centering
  \framebox{\includegraphics[width=.9\linewidth]{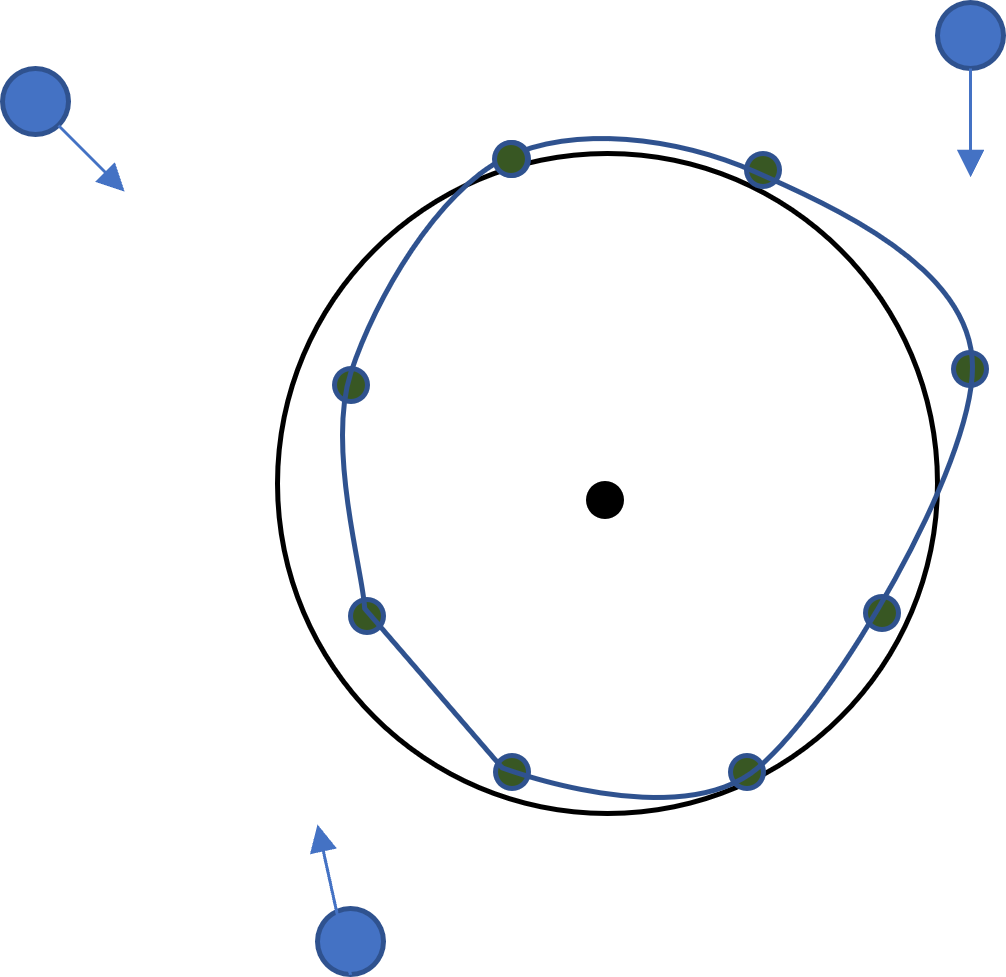}}
  \caption{circular roads}
  \label{fig:roundabout_phase-2}
\end{subfigure}\hspace{0.01\linewidth}%
\begin{subfigure}[b]{.32\linewidth}
  \centering
  \framebox{\includegraphics[width=.9\linewidth]{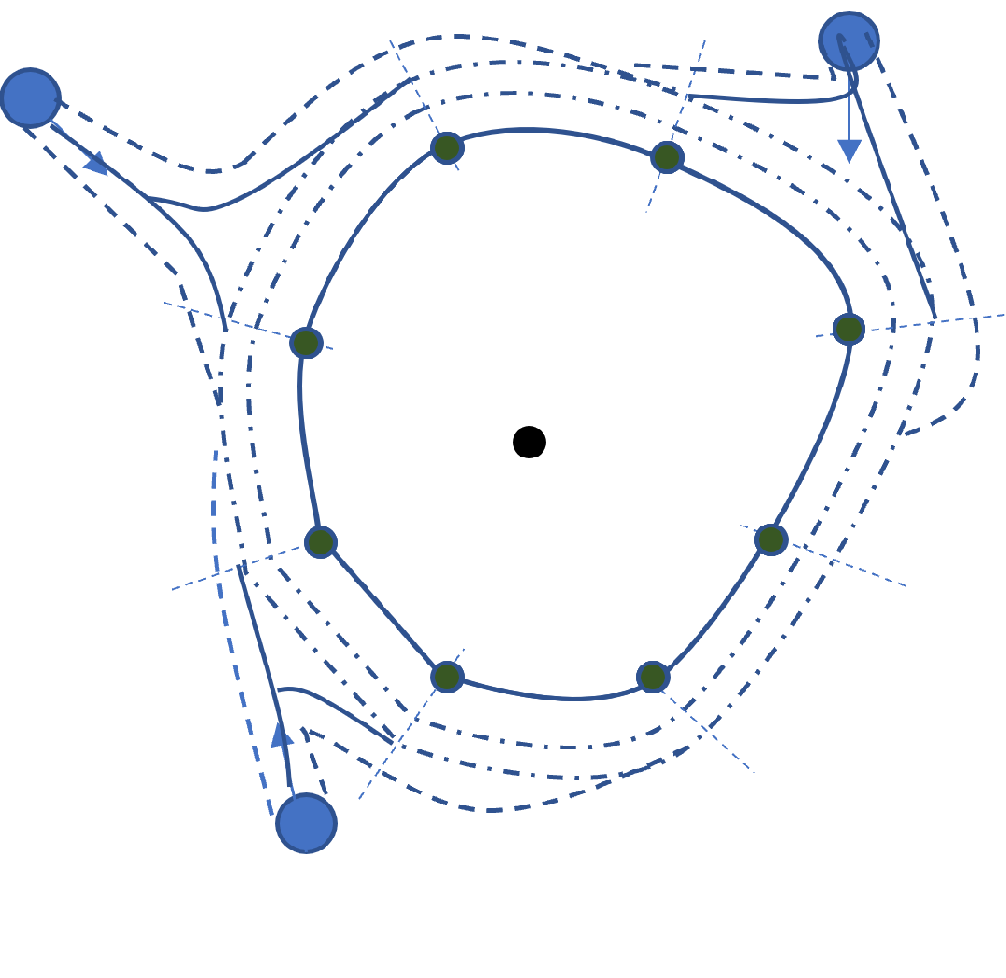}}
  \caption{roundabout}
  \label{fig:roundabout_phase-3}
\end{subfigure}
\newline

\caption{(a) Phase 1 finds maximal circle with input road definitions. (b) Phase 2 builds the skeleton by building  circular roads. (c) Phase 3 adds incident roads to the skeleton built in Phase 2 and finishes the roundabout.}
\label{fig:3-phases}
\end{figure}

%% file: tables-figures/phase2classic.tex
\begin{figure}[ht]
    \begin{subfigure}[b]{.33\linewidth}
      \centering
      \framebox{\includegraphics[width=.9\linewidth]{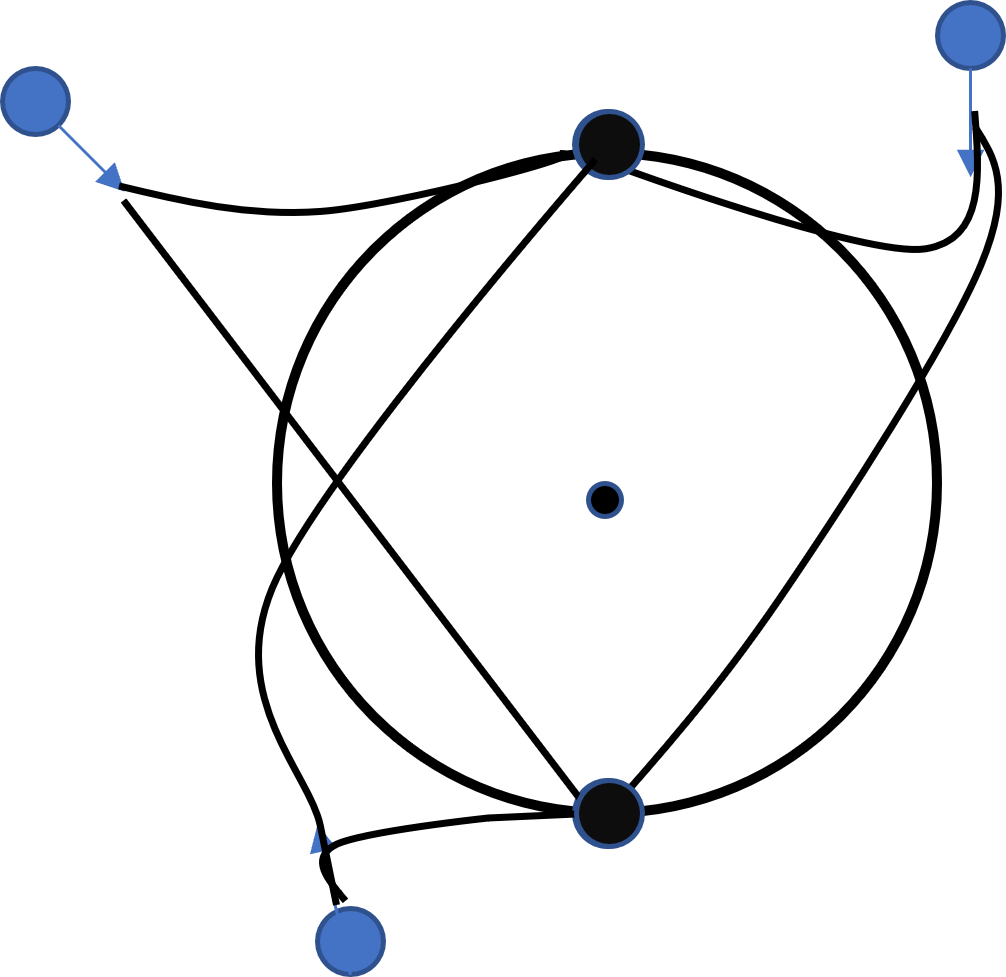}}
      \caption{}
      \label{fig:invalidconnection}
    \end{subfigure}
    \begin{subfigure}[b]{.33\linewidth}
      \centering
      \framebox{\includegraphics[width=.9\linewidth]{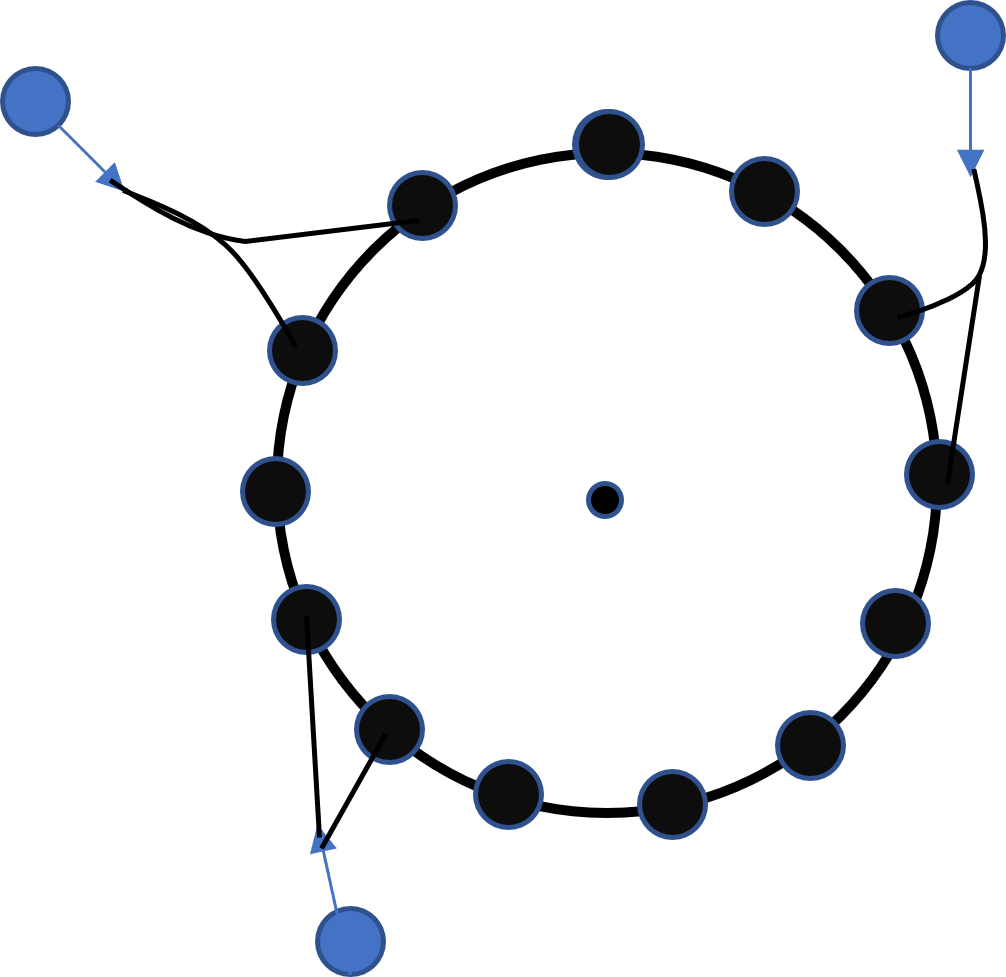}}
      \caption{}
      \label{fig:moresegments}
    \end{subfigure}
    \begin{subfigure}[b]{.33\linewidth}
      \centering
      \framebox{\includegraphics[width=.9\linewidth]{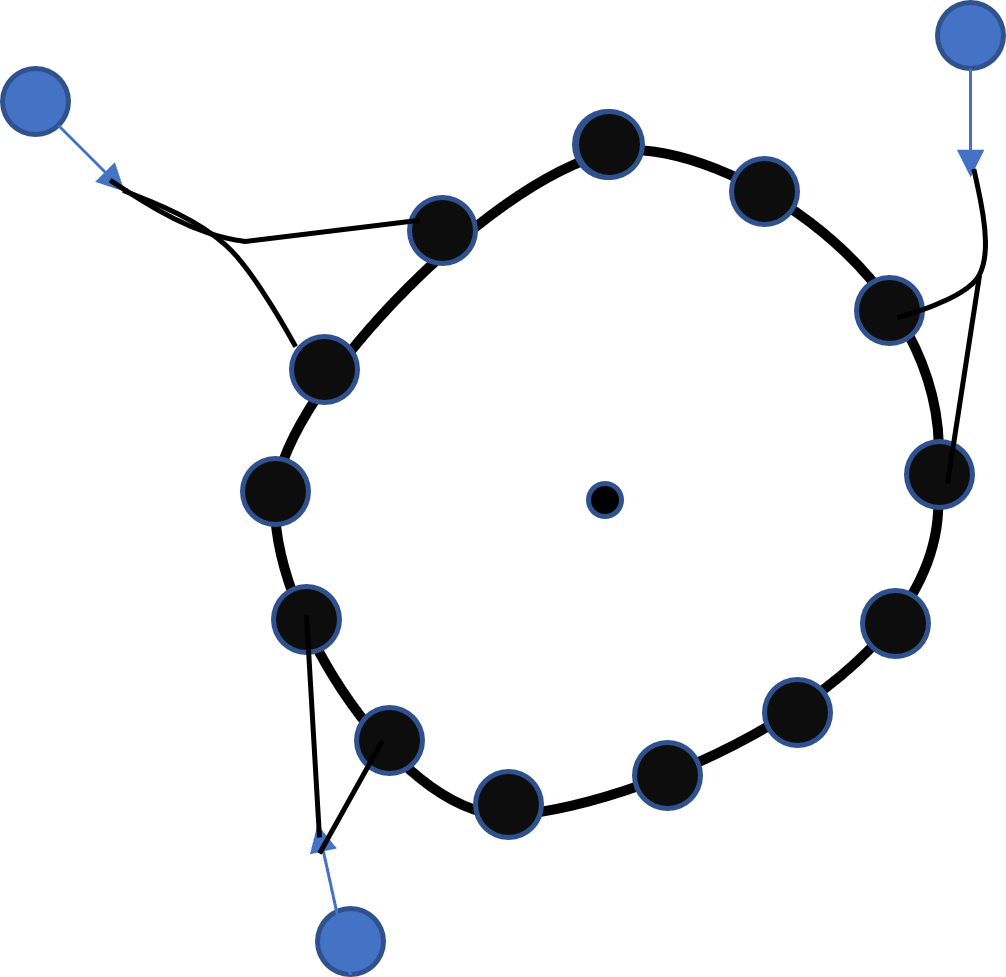}}
      \caption{}
      \label{fig:moresegmentPerlin}
    \end{subfigure}
    \caption{Segmenting the roundabout into small parts for incident road connectivity. Blue filled circles are incident
 points, black arcs are road segments created on the centerline from phase 1, and black filled circle denote the start/endpoints for the segments. (a) Roads can only be linked to start or end points, thus fewer segments lead to road overlap. (b)(c) Dividing the circle into tiny segments allows incident roads to be connected to spatially nearby segments only.}
\end{figure}

%% file: tables-figures/offset_figure.tex
\begin{figure}[ht]
    \begin{subfigure}[b]{.35\linewidth}
      \centering
      \framebox{\includegraphics[width=.9\linewidth]{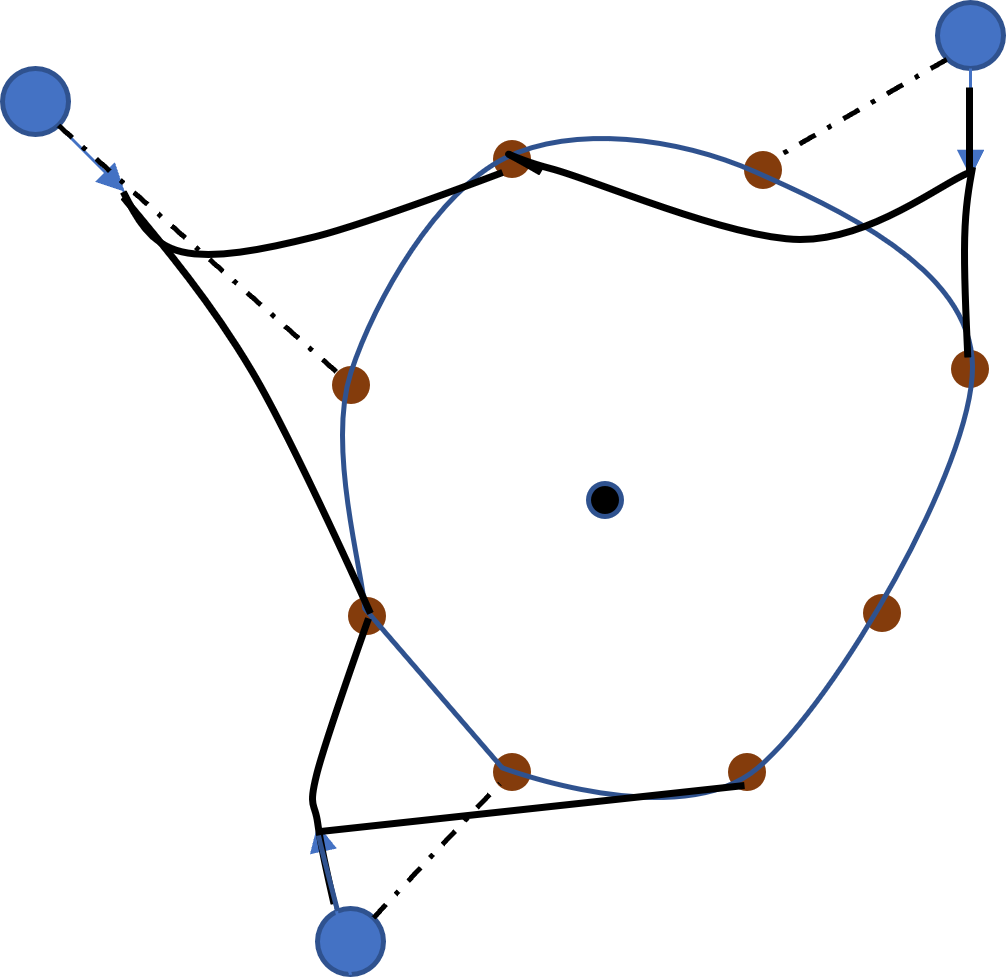}}
      \caption{}
      \label{fig:badconnection}
    \end{subfigure}
    \begin{subfigure}[b]{.65\linewidth}
      \centering
      \framebox{\includegraphics[width=.9\linewidth]{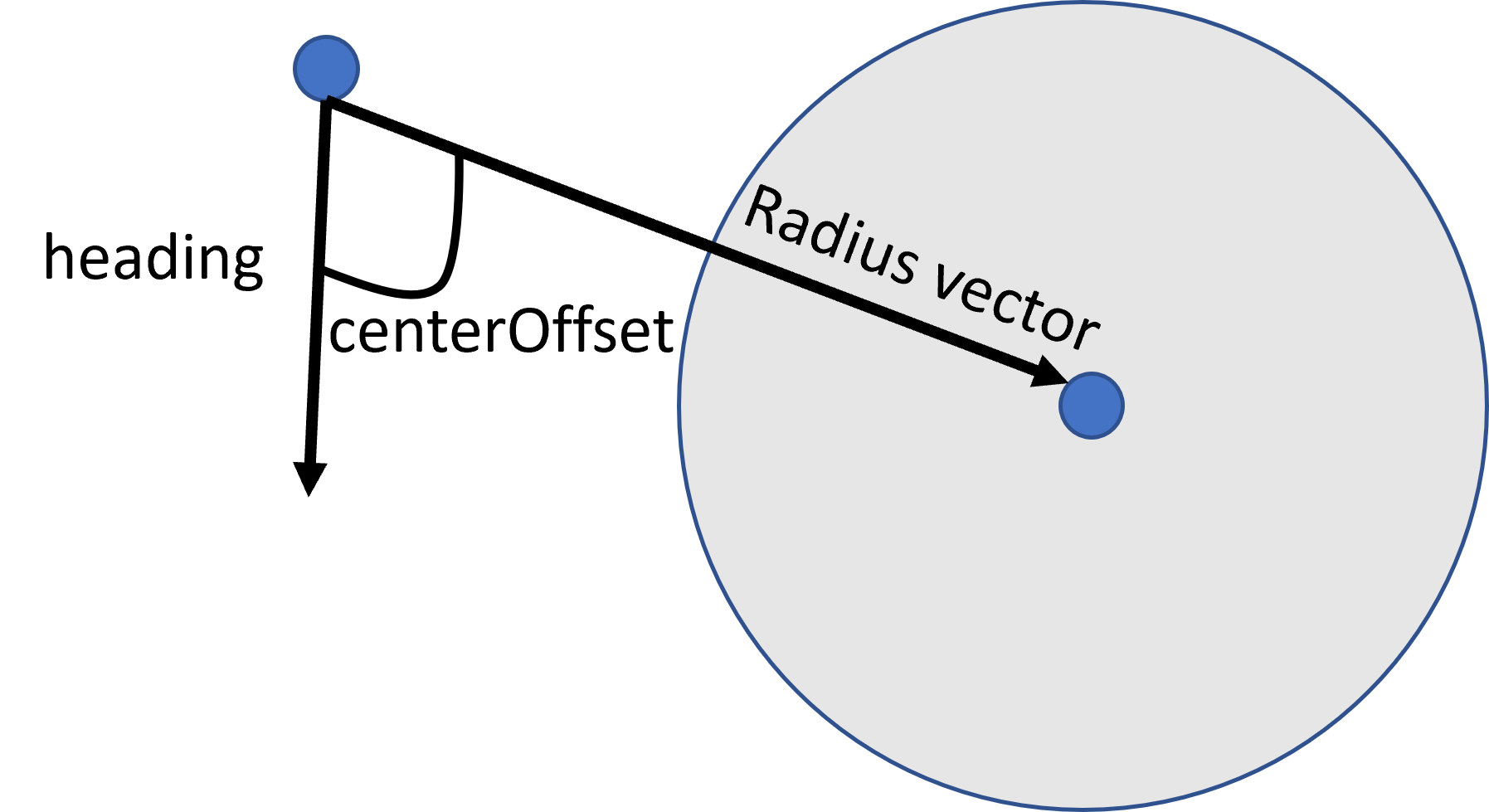}}
      \caption{}
      \label{fig:offset}
    \end{subfigure}
    \caption{(a) fixed connection rule leads to a bad connection (closest segment starting point for each incident points shown with dotted lines), (b) \textit{centerOffset}}.
\end{figure}

%% file: methods/turbo.tex
\subsection{Turbo Roundabout Generation}
In this section we describe the three phases of Turbo Roundabout Generation:

\vspace{5mm}
\subsubsection{\textbf{Phase 1: Finding Maximal Circle Which Has No Incident Points Inside}}

This phase is similar to phase 1 of Classic Roundabout Generation.

\vspace{5mm}
\subsubsection{\textbf{Phase 2: Creating Circular Roads} }

\input{tables-figures/turbo_generation_phase2.tex}
Turbo roundabout circular road structure differs from the classic roundabout because of its spiral pattern. This pattern emerges from circular road segments being translated along one or many axes. For this work we discuss construction of one-translation-axis turbo roundabouts, the most popular variant. Since there is one axis translation, two half circular roads need to be created and translated along the chosen axis. Besides, in the intersection where half circular roads are translated, incident roads are connected differently as the incoming and outgoing incident lanes connect to two structures that belong to different circles. To capture these behaviors, we define \textit{spike} as the road segments that connect the two half circular segments. Phase 2 has the following steps:
\begin{enumerate}
    \item Find a pair of optimal points which will characterize the translation axis for spike generation (Fig. \ref{fig:turbo-step-1},\ref{fig:turbo-step-2}).
    \item Using radius and center, create two half circular roads using multiple segments, translate them and connect them using straight roads (spike). (Fig. \ref{fig:turbo-step-3},\ref{fig:turbo-step-4})
    \item Rotate road system to the translation axis. (Fig. \ref{fig:turbo-step-5})
\end{enumerate}

\vspace{5mm}
\subsubsection{\textbf{Phase 3 : Creating Incident Roads}}
\input{tables-figures/turbo_generation_phase3.tex}
The only difference between this phase and that of the Classic Roundabout Generation is that in this phase, compatible points will be connected to spike (Fig:\ref{fig:turbo_step6}), so that the roads generated from it are smoothly connected (Fig:\ref{fig:turbo_step7}), resembling a turbo roundabout. We connect the rest of the incident points using \textit{centerOffset} as shown in the Phase 3 of Classic Roundabout Generation.

%% file: tables-figures/turbo_generation_phase2.tex
\begin{figure}[ht]
    \centering
    \begin{subfigure}[b]{.31\linewidth}
      \centering
      \framebox{\includegraphics[width=.87\linewidth]{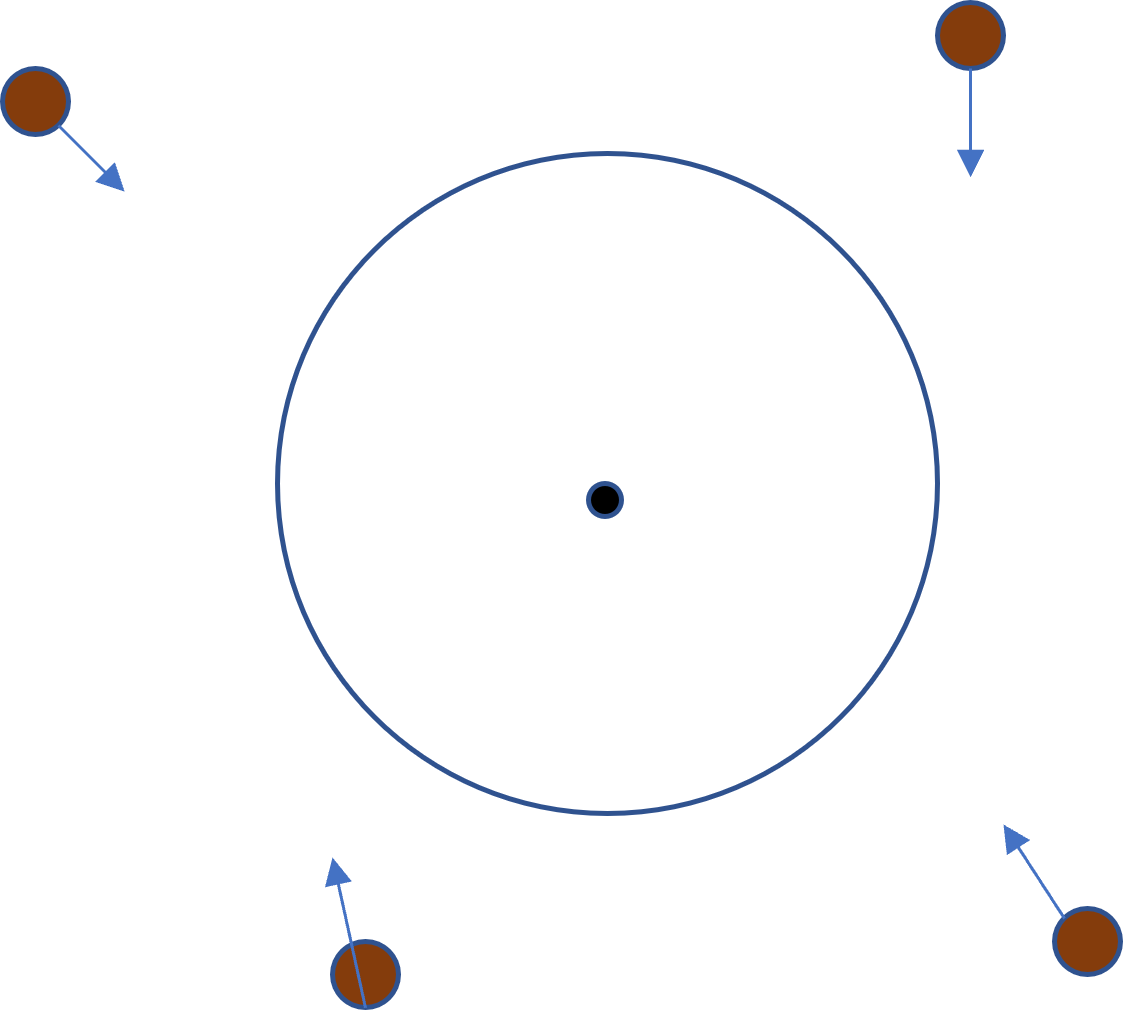}}
      \caption{}
      \label{fig:turbo-step-1}
    \end{subfigure}\hspace{0.03\linewidth}%
    \begin{subfigure}[b]{.31\linewidth}
      \centering
      \framebox{\includegraphics[width=.9\linewidth]{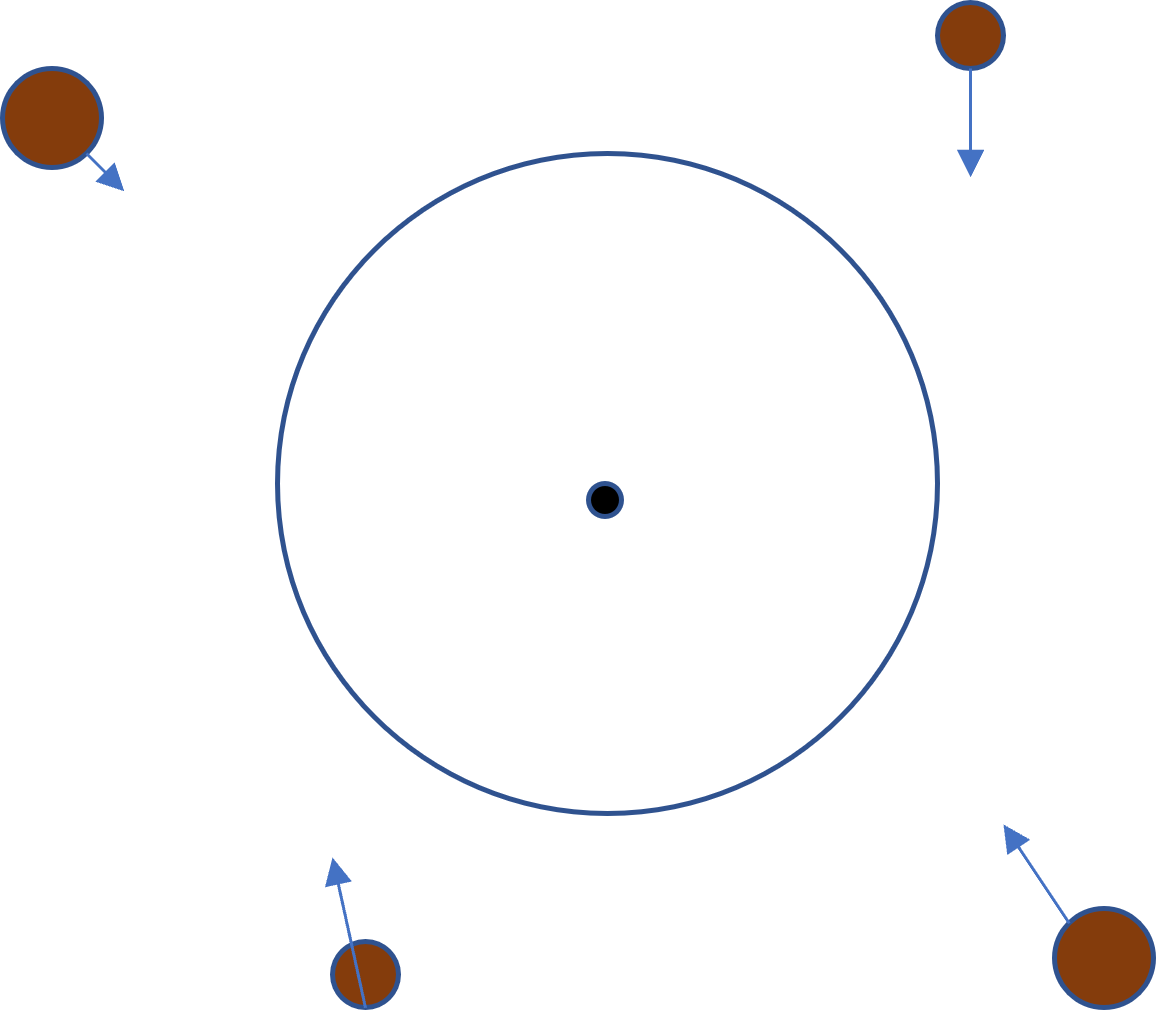}}
      \caption{}
      \label{fig:turbo-step-2}
    \end{subfigure}\hspace{0.03\linewidth}%
    \begin{subfigure}[b]{.31\linewidth}
      \centering
      \framebox{\includegraphics[width=.9\linewidth]{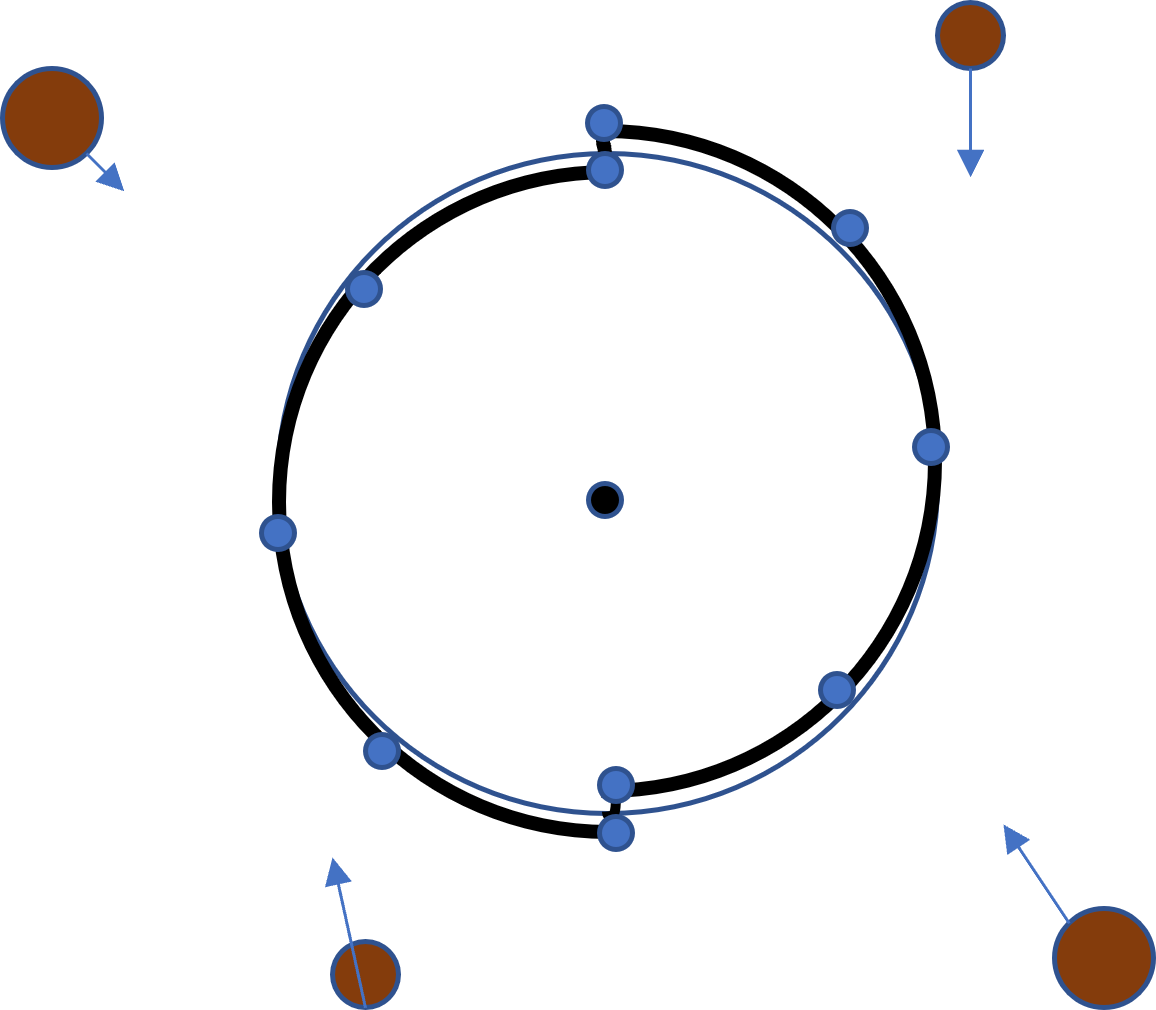}}
      \caption{}
      \label{fig:turbo-step-3}
    \end{subfigure}
    \newline
    \begin{subfigure}[b]{.31\linewidth}
      \centering
      \framebox{\includegraphics[width=.9\linewidth]{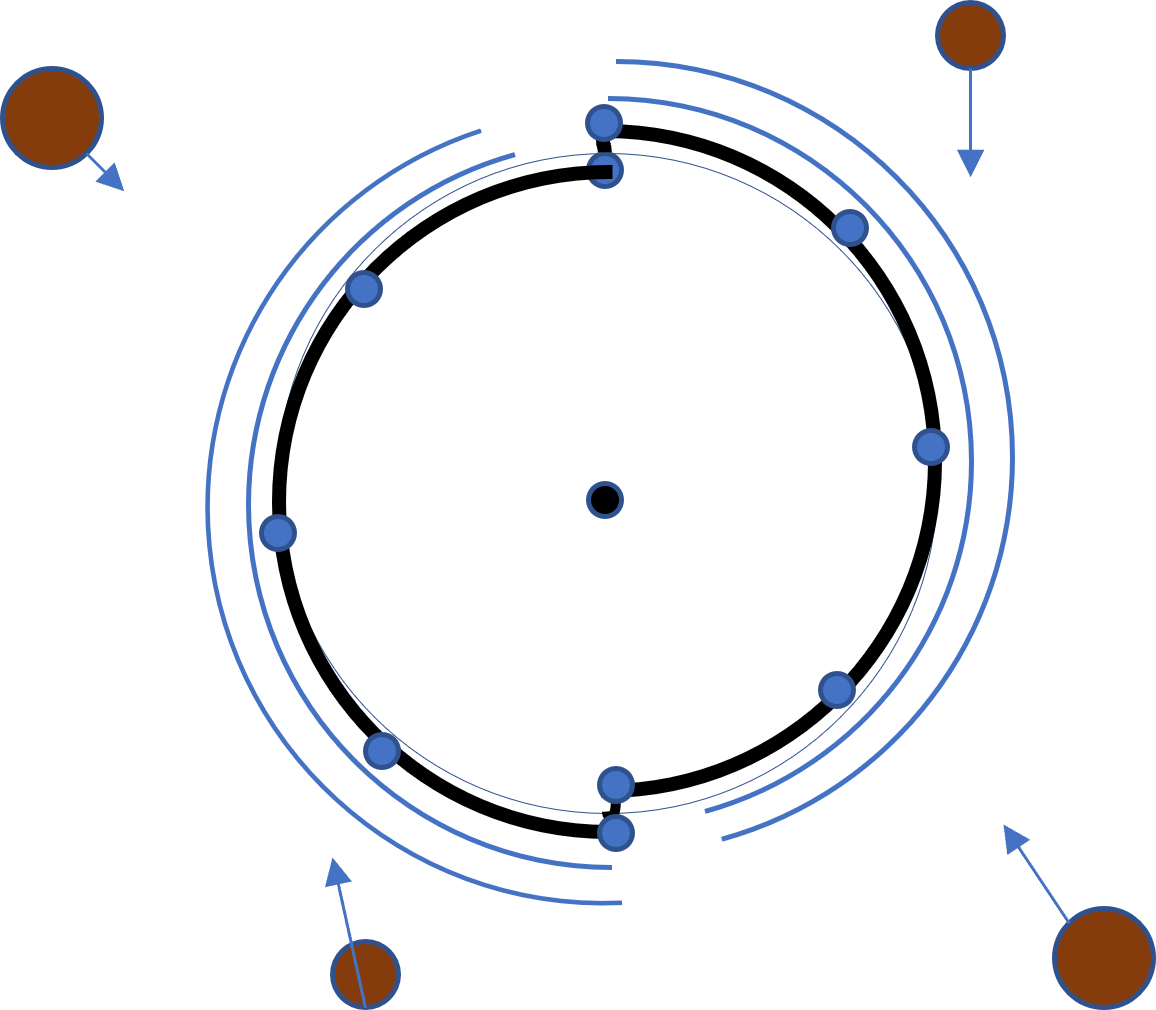}}
      \caption{}
      \label{fig:turbo-step-4}
    \end{subfigure}%
    \begin{subfigure}[b]{.31\linewidth}
      \centering
      \framebox{\includegraphics[width=.9\linewidth]{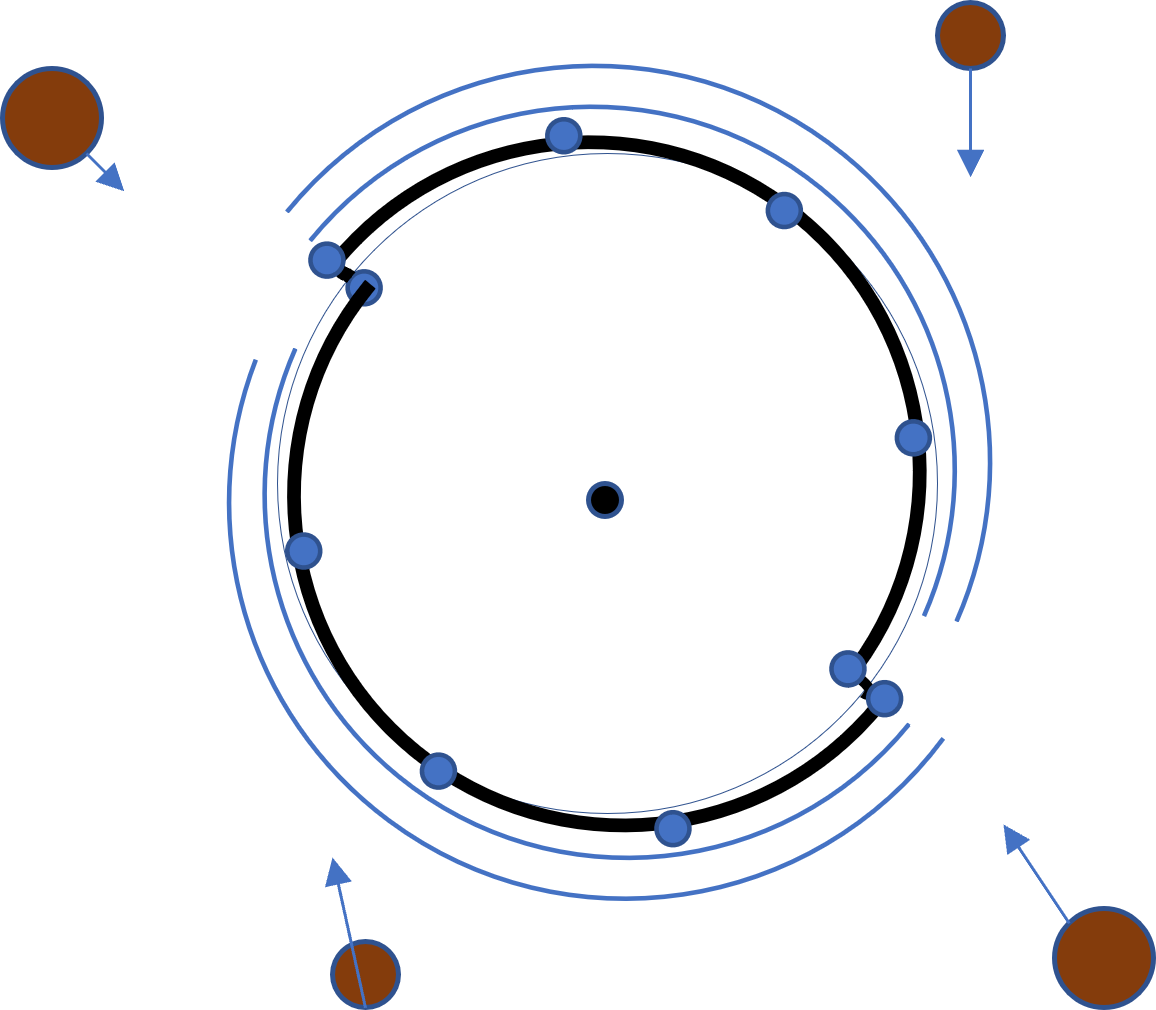}}
      \caption{}
      \label{fig:turbo-step-5}
    \end{subfigure}
    \caption{(a)(b) Finding compatible point (compatible point shown with bigger radius), (c)(d) Creating Turbo Circular Roads, (e) Rotating structure to match compatible points}
    \label{fig:turbo-phase2}
\end{figure}

%% file: tables-figures/turbo_generation_phase3.tex
\begin{figure}[bt!]
    \centering
    \begin{subfigure}[b]{.3\linewidth}
      \centering
      \framebox{\includegraphics[width=.9\linewidth]{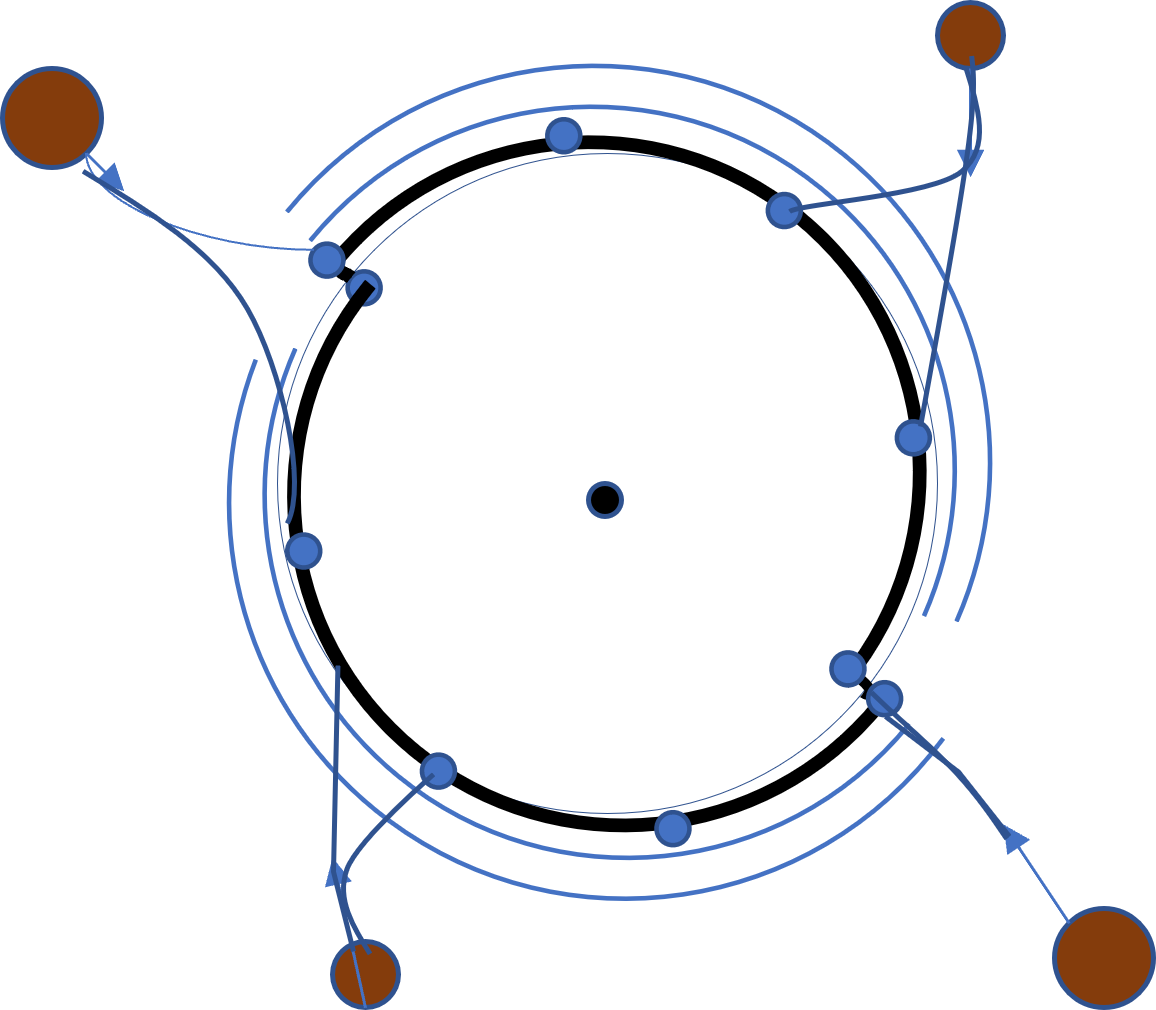}}
      \caption{}
      \label{fig:turbo_step6}
    \end{subfigure}
    \begin{subfigure}[b]{.3\linewidth}
      \centering
      \framebox{\includegraphics[width=.9\linewidth]{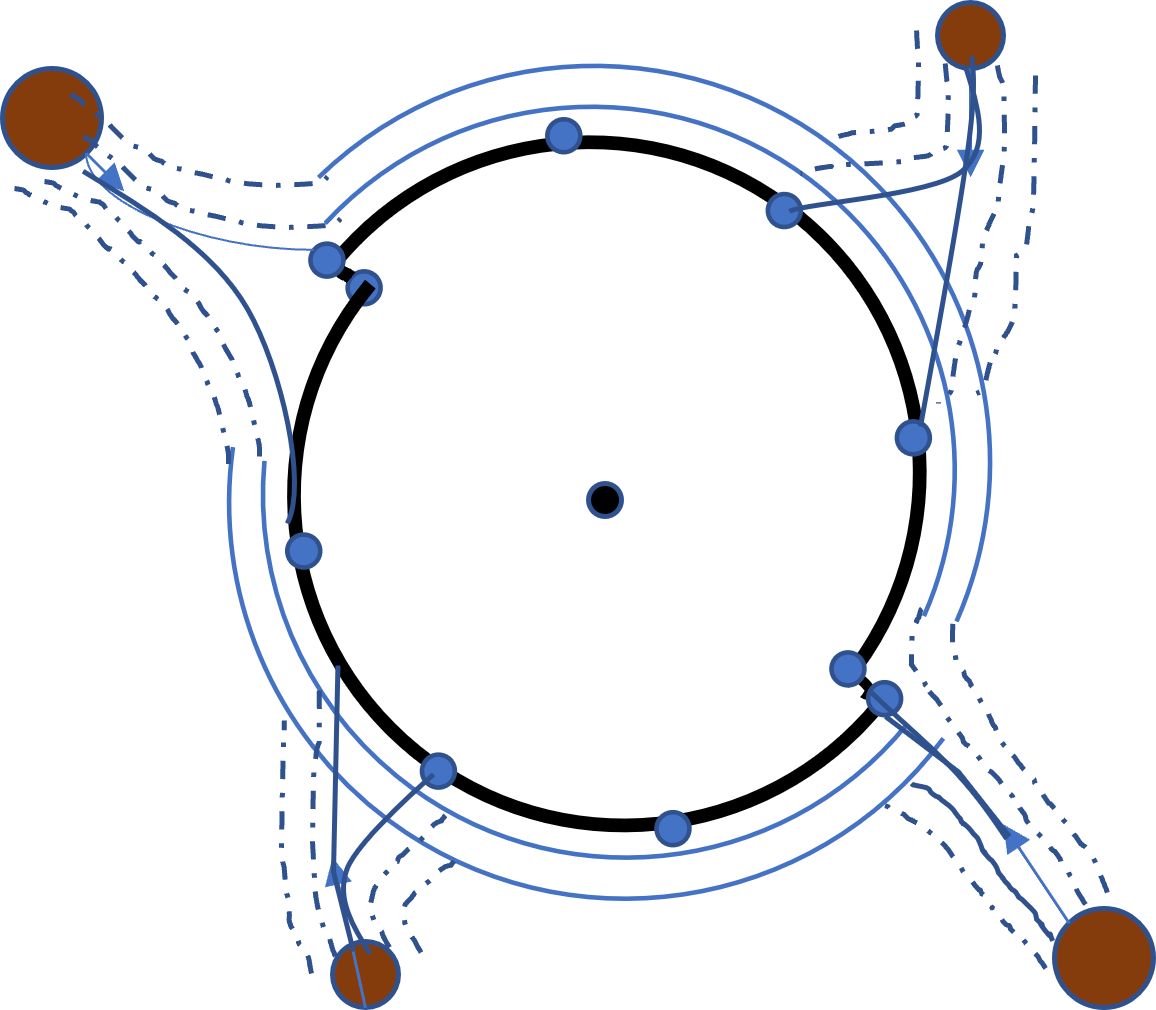}}
      \caption{}
      \label{fig:turbo_step7}
    \end{subfigure}
    \caption{(a) Connecting compatible points to the spike and other points to opmimum locations, (b) Adding connection roads.}
    \label{fig:turbo_phase3}
\end{figure}

%% file: tables-figures/beautifulroundabouts.tex
\begin{figure}[bt!]
    \begin{subfigure}[b]{.33\linewidth}
      \centering
      \framebox{\includegraphics[width=.9\linewidth]{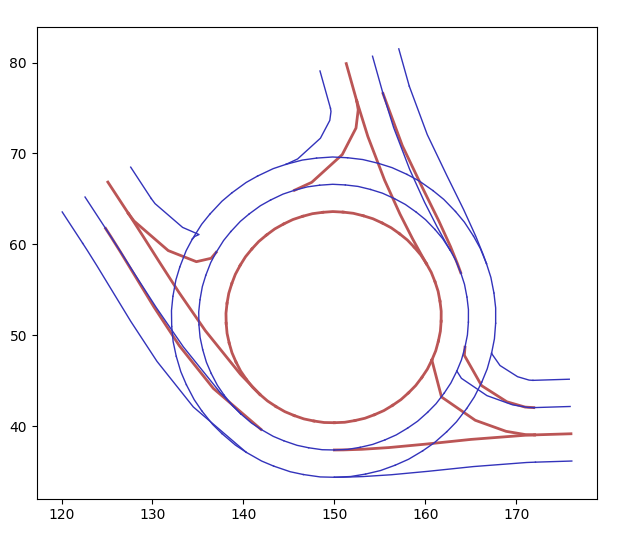}}
      \caption{}
      \label{r1}
    \end{subfigure}
    \begin{subfigure}[b]{.3\linewidth}
      \centering
      \framebox{\includegraphics[width=.9\linewidth]{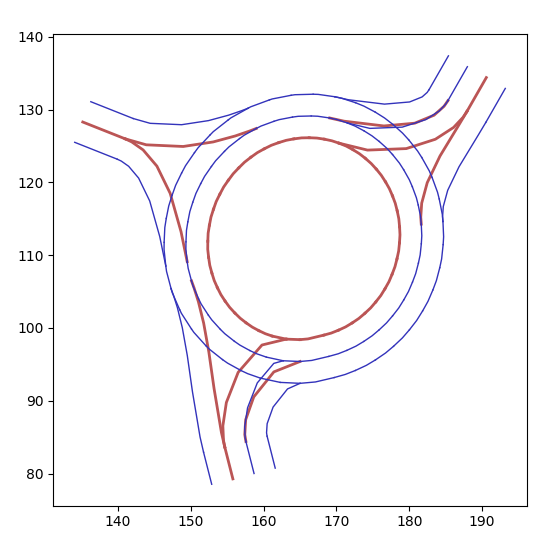}}
      \caption{}
      \label{r2}
    \end{subfigure}
    \begin{subfigure}[b]{.34\linewidth}
      \centering
      \framebox{\includegraphics[width=.9\linewidth]{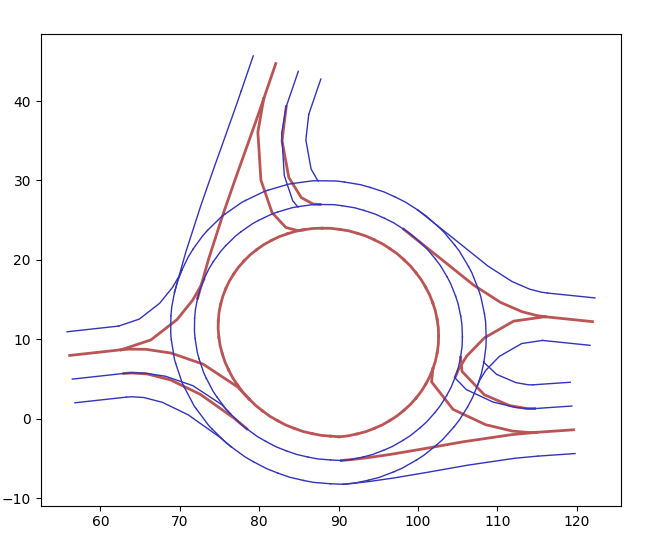}}
      \caption{}
      \label{r3}
    \end{subfigure}
    \caption{Randomly generated classic roundabouts. In (a) we have three-lane incident roads, in (b) and (c) the incident lanes have either two lanes or three lanes.}
    \label{fig:classicroundabouts}
\end{figure}

%% file: tables-figures/beautifulturboroundabouts.tex
\begin{figure}[bt!]
    \begin{subfigure}[b]{.43\linewidth}
      \centering
      \framebox{\includegraphics[width=.9\linewidth]{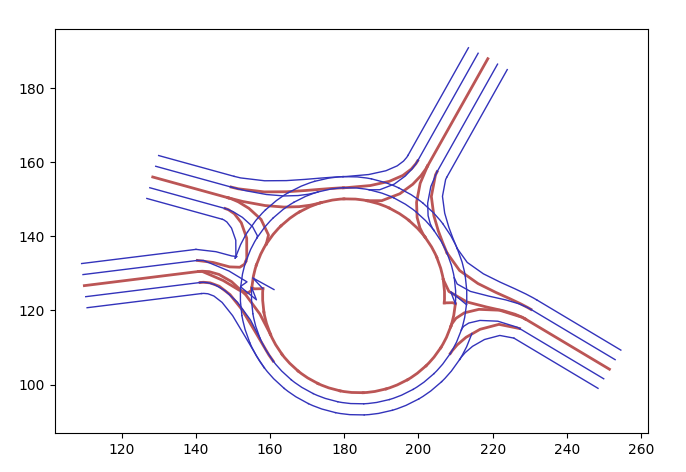}}
      \caption{}
      \label{tr1}
    \end{subfigure}
    \begin{subfigure}[b]{.24\linewidth}
      \centering
      \framebox{\includegraphics[width=.9\linewidth]{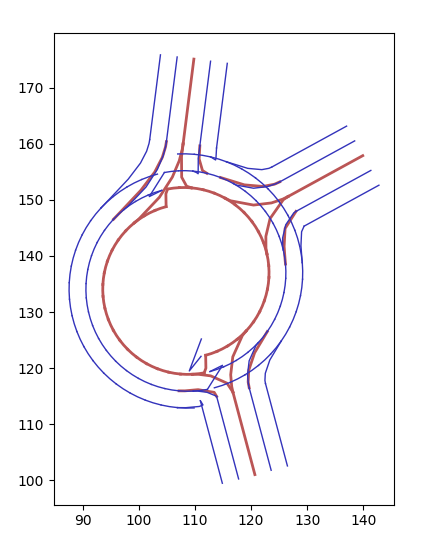}}
      \caption{}
      \label{tr2}
    \end{subfigure}
    \begin{subfigure}[b]{.31\linewidth}
      \centering
      \framebox{\includegraphics[width=.9\linewidth]{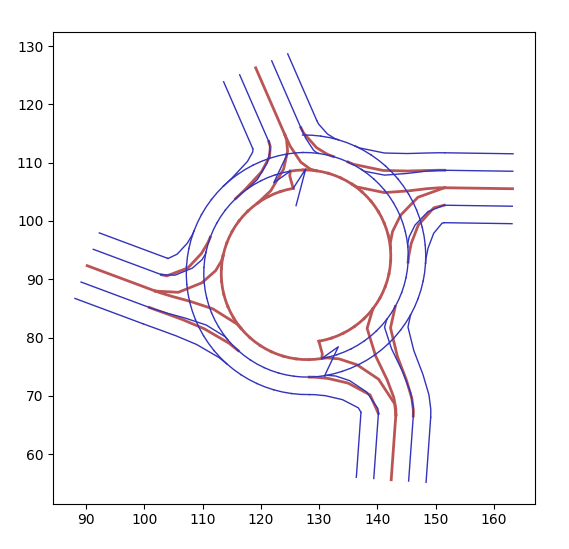}}
      \caption{}
      \label{tr3}
    \end{subfigure}
    \caption{Randomly generated turbo roundabouts.}
    \label{fig:turboroundabouts}
\end{figure}

%% file: tables-figures/comparison.tex
\begin{figure}[ht]
    \centering
    \framebox{\includegraphics[width=.9\linewidth]{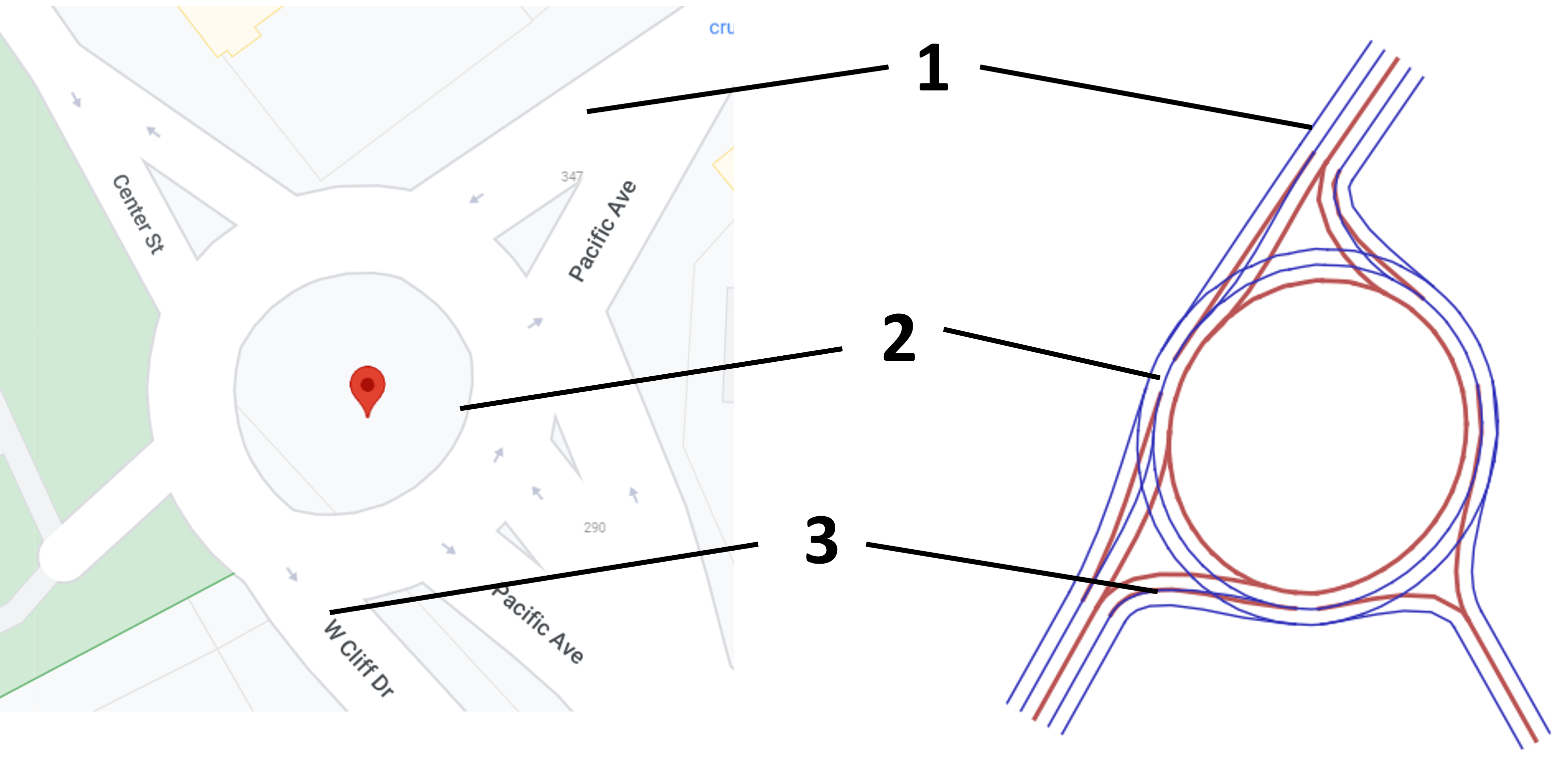}}
    \caption{Comparison of a real-world roundabout from 290 Pacific Ave in Santa Cruz, California and Junction-Art generated output. Feature 1 shows variable incident road lanes, Feature 2 shows semi circular shape, and Feature 3 shows variable incident road angles. Map data \textcopyright 2023 Google}
    \label{fig:comparsion}
\end{figure}

%% file: tables-figures/pointgen.tex
\begin{figure}[ht!]

    \centering
    \framebox{\includegraphics[width=.9\linewidth]{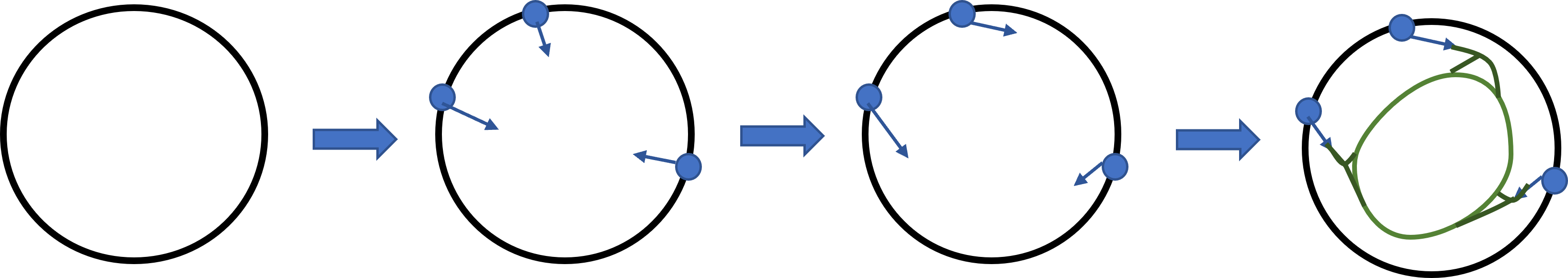}}
    \caption{Random road definition generation process for evalution.}
    \label{fig:roaddef}
\end{figure}

%% file: tables-figures/circlevariation.tex
\begin{figure}[!ht]

    \centering
    \includegraphics[width=.6\linewidth]{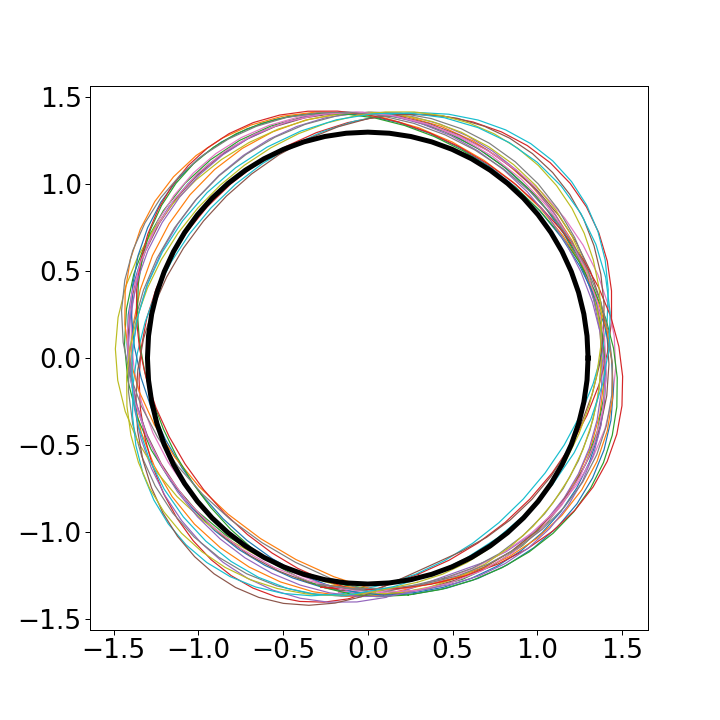}
    \caption{Normalized 3-way roundabouts superimposed on a circle. Thick black line resembles the circle.}
    \label{fig:circlecomp}
\end{figure}

%% file: tables-figures/radiusdist.tex
\begin{figure}[bt!]
    \begin{subfigure}[b]{\linewidth}
      \centering
      \includegraphics[width=\linewidth]{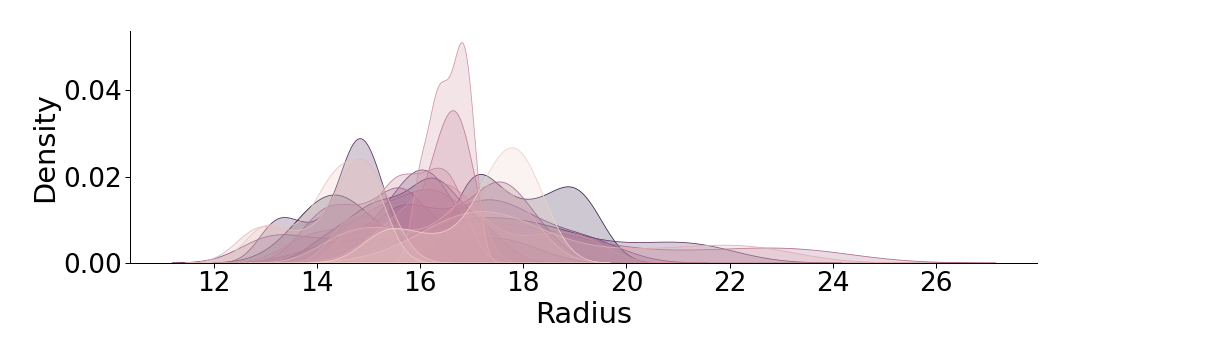}
      \caption{}
      \label{fig:3-way}
    \end{subfigure}
    \begin{subfigure}[b]{\linewidth}
      \centering
      \includegraphics[width=\linewidth]{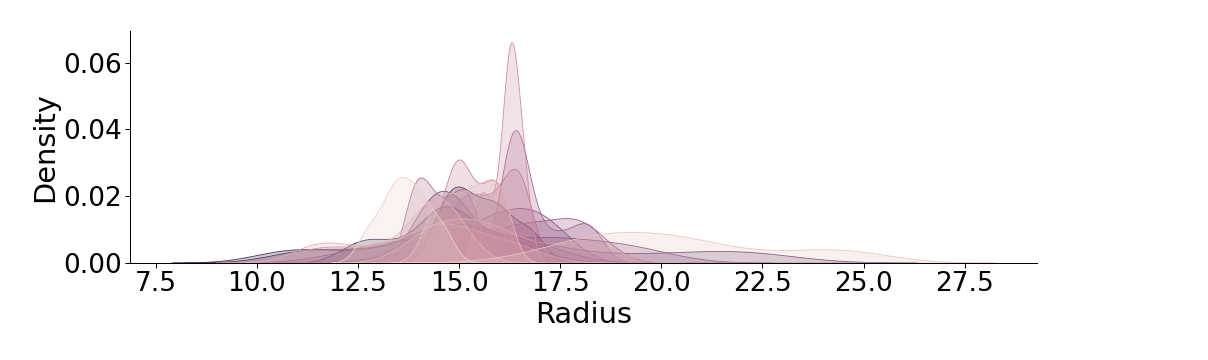}
      \caption{}
      \label{fig:4-way}
    \end{subfigure}
    \begin{subfigure}[b]{\linewidth}
      \centering
      \includegraphics[width=\linewidth]{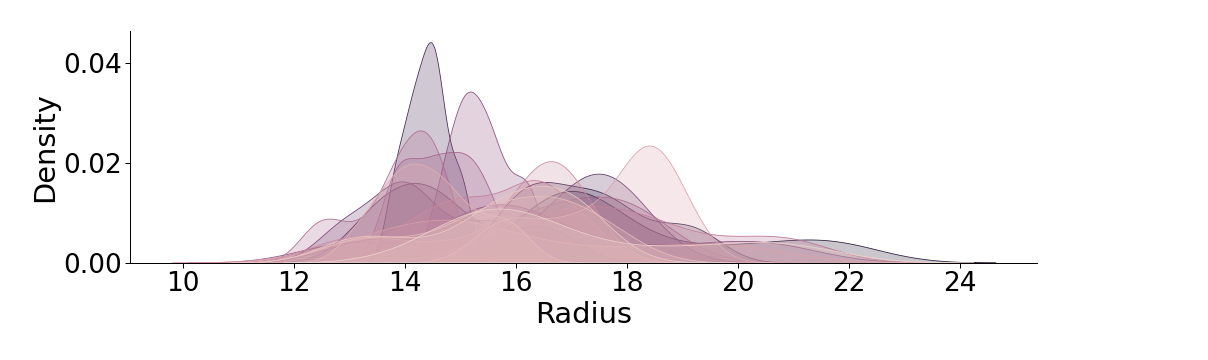}
      \caption{}
      \label{fig:5-way}
    \end{subfigure}
    \caption{Distribution of radii in several (a) 3, (b) 4, (c) 5 ways roundabouts. Each line represents a roundabout's radius at different points on its center line. Driving on a roundabout with wider spread is harder in general.}
    \label{fig:radiusDis}
\end{figure}

%% file: tables-figures/derivative.tex
\begin{figure}[ht!]

    \centering
    \includegraphics[width=\linewidth]{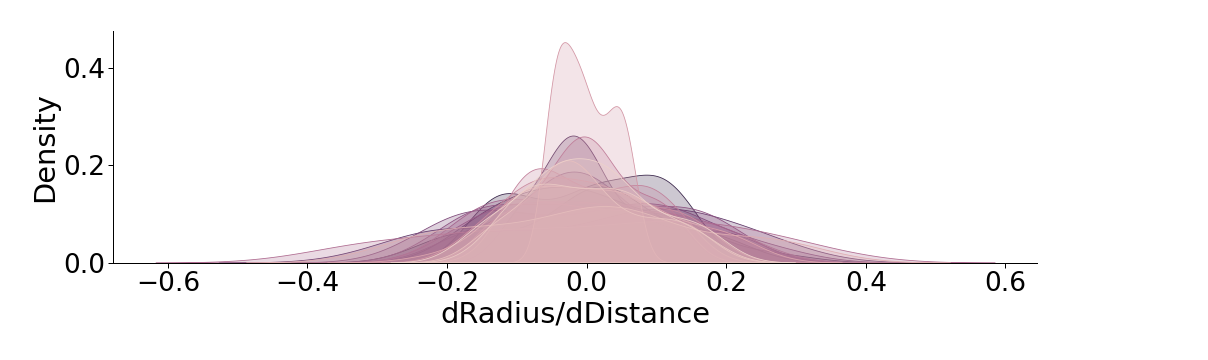}
    \caption{Distribution of the derivative of radii in several 3-way roundabouts.}
    \label{fig:derivative}
\end{figure}

%% file: tables-figures/3wayfixedradius.tex
\begin{figure}[bt!]

    \centering
    \includegraphics[width=.9\linewidth]{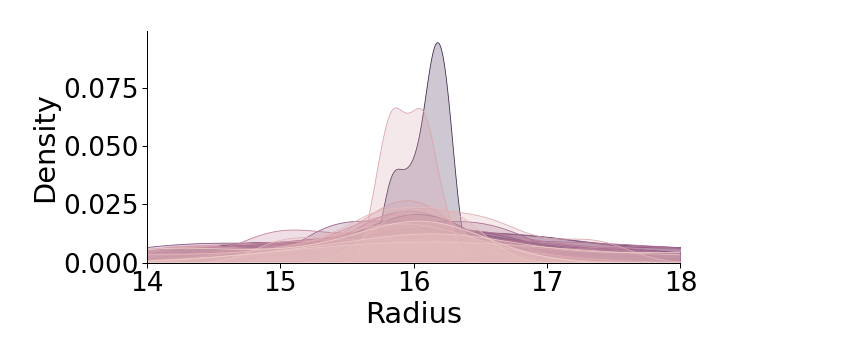}
    \caption{Distribution of radii in several 3-way roundabouts generated from the same road definition. Each line represents a roundabout's radius at different points on its center line.}
    \label{fig:fixedRadDist}
\end{figure}

%% file: tables-figures/fixed_roundabout_dis.tex
\begin{figure}[bt!]
    \begin{subfigure}[b]{.5\linewidth}
      \centering
      \includegraphics[width=\linewidth]{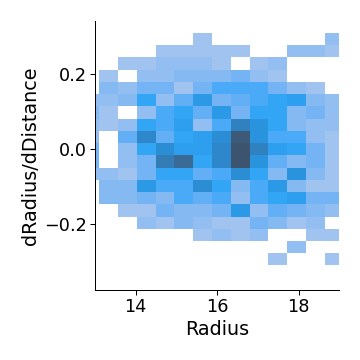}
      \caption{}
      \label{fig:fixed_dist}
    \end{subfigure}
    \begin{subfigure}[b]{.5\linewidth}
      \centering
      \includegraphics[width=\linewidth]{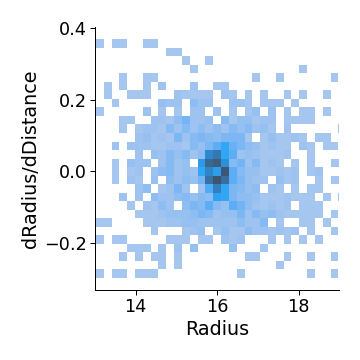}
      \caption{}    
      \label{fig:fixed_heatmap}
    \end{subfigure}
    \caption{Bivariate density estimation of radii and derivative of the radii in several 3-way roundabouts generated for (a)random road definitions (b) a fixed road definition.}

\end{figure}